\def\year{2020}\relax
\documentclass[letterpaper]{article} 
\usepackage{aaai20}  
\usepackage{times}  
\usepackage{helvet} 
\usepackage{courier}  
\usepackage[hyphens]{url}  
\usepackage{graphicx} 
\usepackage{graphicx}  
\usepackage{amsmath}
\usepackage{amssymb}
\usepackage{booktabs} 
\usepackage{multirow}

\newcommand{\eg}{\textit{e}.\textit{g}.}

\title{Accurate Temporal Action Proposal Generation with \\Relation-Aware Pyramid Network}

\author{Jialin Gao\textsuperscript{\rm $1$}, Zhixiang Shi\textsuperscript{\rm 2}, Jiani Li\textsuperscript{\rm 2}, Guanshuo Wang\textsuperscript{\rm 1},\\ \bf \Large Yufeng Yuan\textsuperscript{\rm 2}, Shiming Ge\textsuperscript{\rm 3}\thanks{Shiming Ge is the corresponding author.}, and Xi Zhou\textsuperscript{\rm 1,2}  \\
	\textsuperscript{\rm 1}Cooperative Medianet Innovation Center, Shanghai Jiao Tong University \\
	\textsuperscript{\rm 2}CloudWalk Technology Co., Ltd, China \\
	\textsuperscript{\rm 3} Institute of Information Engineering, Chinese Academy of Sciences\\
	\textsuperscript{}\{shizhixiang, lijiani, yuanyufeng, zhouxi\}@cloudwalk.cn,
	\{jialin\_gao,guanshuo.wang\}@sjtu.edu.cn,~geshiming@iie.ac.cn}

\begin{document}
	
	\maketitle
	
	\begin{abstract}
		Accurate temporal action proposals play an important role in detecting actions from untrimmed videos. The existing approaches have difficulties in capturing global contextual information and simultaneously localizing actions with different durations. To this end, we propose a Relation-aware pyramid Network (RapNet) to generate highly accurate temporal action proposals. In RapNet, a novel relation-aware module is introduced to exploit bi-directional long-range relations between local features for context distilling. This embedded module enhances the RapNet in terms of its multi-granularity temporal proposal generation ability, given predefined anchor boxes. We further introduce a two-stage adjustment scheme to refine the proposal boundaries and measure their confidence in containing an action with snippet-level actionness. Extensive experiments on the challenging ActivityNet and THUMOS14 benchmarks demonstrate our RapNet generates superior accurate proposals over the existing state-of-the-art methods.
	\end{abstract}
	
	\section{Introduction}
	Temporal action proposal generation aims to propose candidate instances that probably contain an action from a long, untrimmed video. Localizing human activities is of increasing importance as it brings about a wide variety of application in industry and serves as the basis in many research areas, like action detection\cite{zhao2017temporal}, video summarization \cite{gong2014diverse,yao2016highlight}, captioning \cite{wang2018bidirectional,wang2018reconstruction}, etc.
	
	Current methods are mainly divided into two categories: sliding-windows and snippet-level actionness score. The former \cite{shou2016temporal,shou2017cdc} slides several manually defined windows to generate proposals with imprecise temporal boundaries and rank them via a binary classifier. Instead, the latter \cite{xiong2017pursuit,yuan2017temporal} generates candidate instances with high precision boundaries by actionness score grouping \cite{zhao2017temporal}. But it has incorrect and omitted candidates due to the difficulty in identifying action without long-term contextual information.
	
	\begin{figure}[t]
		\centering
		\includegraphics[width=0.95\columnwidth]{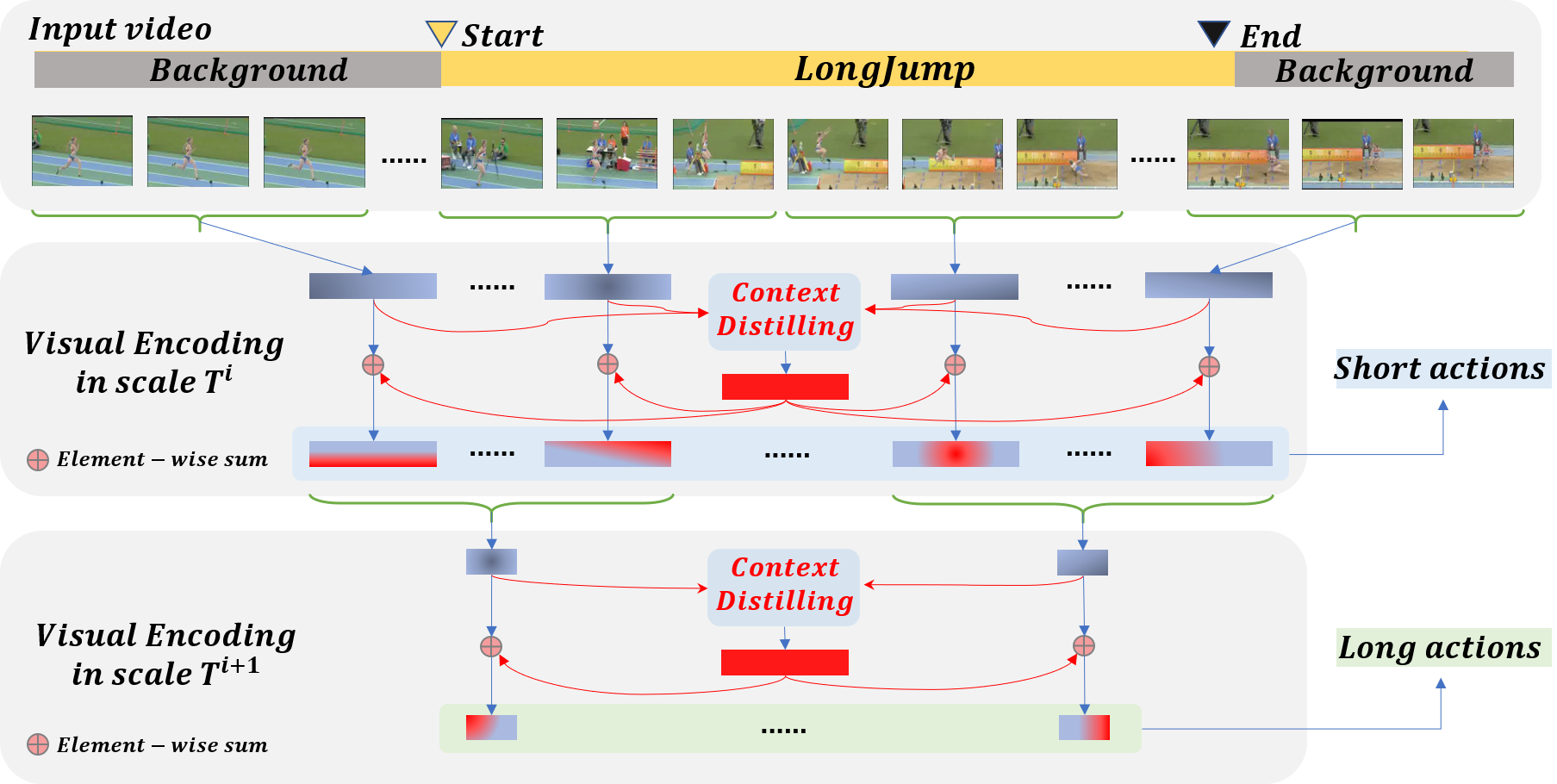}
		\caption{For action localization, two crucial issues lie in (a) how to simultaneously localize action instances with various durations, and (b) how to distill long-range context to capture informative contents for high precision boundaries. The former can be solved by pyramid prediction and the latter needs a novel context distilling module.}
		\label{motivation}
	\end{figure}
	
	Several attempts have been made to overcome these drawbacks. Boundary sensitive network (BSN) \cite{lin2018bsn} identifies each snippet to be either starting, ending or course for generating proposals with various durations. Complementary Temporal Action Proposal (CTAP) \cite{gao2018ctap} alleviates the problem of missing contextual information by combining the results with window-based methods. However, they only consider the complementary characteristics of these two types of methods, while do not address their intrinsic difficulties.

	In order to address difficulties in accurately localizing actions, we refer to the perception mechanism of human. Considering a complete action may last from a second to several minutes, humans prefer to skim though long instances quickly while oversee short ones carefully in order to focus on informative contents rather than backgrounds. To accurately discriminate the starting and ending of an action, we humans usually relate the present to the long-range contextual information. These reveal two key factors relating to accurate temporal proposal generation: \emph{how to simultaneously detect actions in multi-scales} and \emph{how to distill long-range context}. As shown in Fig.\ref{motivation}, the context distilling block is required to extract long-range contextual information for augmenting corresponding local features, which are responsible for generating multi-granularity action proposals.

	For accurate action instance localization, the first factor could guarantee the generated proposals adaptive to various duration of actions, and the second factor is introduced to distinguish action contents from backgrounds for high precision boundary. To this end, a proposal generation approach is required to capture long-range contextual information at different temporal scales and then augment the corresponding local features. Previous bottom-up methods, although useful in locating action boundaries and prediction actionness score, are based only on local information and ignore the long-range contextual information.

	Therefore, a Relation-aware-pyramid Network (RapNet) is proposed to generate accurate temporal action proposals. We firstly introduce a novel Relation-Aware Module (RAM) to exploit long-range contextual information, which is ignored in previous methods. This module captures bi-directional relations between local patterns as context distilling, in order to relate the present local content to both the past and future. Temporal information is squeezed within the module to generate channel-wise mean statistics as a global context for augmenting the corresponding local information.
	
	Afterward, by embedding this module, the RapNet enhances its ability to generate multi-scales temporal action proposals with high precision boundaries, given predefined anchor boxes by K-means clustering. Specifically, a global context extractor employs a RAM to boost local features with holistic information. Then a U-shape net with skip connections is developed to construct a bottom-up feature pyramid set, which is used to generate candidates in multi temporal scales via separate generators.
	
	Moreover, considering the precise boundaries of proposal generated by actionness score grouping, we also predict snippet-level actionness score to adjust the boundaries of proposal candidates from RapNet and measure their confidence of containing an action in terms of proposal-level feature.
	
	Finally, we conduct extensive experiments on the challenging THUMOS-14 and ActivityNet-v1.3 benchmarks to demonstrate the effectiveness of our RapNet and relation-aware module. In summary, our contributions are three-fold:
	\begin{itemize}
		\item[1)] We propose a novel context distilling mechanism enhanced with pyramid networks to accurately generate multi-granularity proposals, which captures long-range contextual information at different temporal scales to augment local features.
		\item[2)] We propose the relation-aware module, as an exemplar of context distilling, to exploit bidirectional relations between any two temporal locations, which relates the present content to both the past and future for augmenting multi-granularity temporal proposal generation.
		\item[3)] Our RapNet achieves the state-of-the-art performance on the THUMOS-14 and ActivityNet-v1.3 datasets for the temporal action proposal task and surpasses other methods by a considerable margin.
	\end{itemize}
	
	\section{Related Work}
	\textbf{Temporal Action Proposal Generation.} There are two types of methods for this task, which are sliding windows and actionness score. SCNN-prop \cite{shou2016temporal} regards this task as training a C3D network for binary classification task when generating segmental proposals. For generating precise candidates, TAG \cite{zhao2017temporal}, serving as a typical actionness-based method, adopts watershed algorithm to group contiguous high-score as proposals. BSN \cite{lin2018bsn} additionally predicts whether each temporal point is the starting or ending point of an action and evaluates proposals based on proposal-level feature. CTAP \cite{gao2018ctap} takes advantage of both segment proposal and snippet-level actionness based methods to reorganize the candidate set. MGG \cite{liu2019multi} embeds the temporal location information and employs a bilinear matching module. However, these approaches are based on local information and ignore the global context, which is expected to be useful in accurately locating action boundaries and predicting actionness score.
	
	\textbf{Action Recognition and detection.} Action recognition is the primary issue of video understanding and has been extensively studied. Recent, long-range dependencies is of central importance. Non-local network \cite{wang2018non} introduces traditional non-local algorithm into CNNs. LFB \cite{wu2019long} designs a long-term feature bank to improve state-of-the-art video models. These findings inspire us the importance of global contextual information and long-range dependency in video understanding.
	
	Action detection focuses on detecting what the activity is and when it starts and ends in untrimmed video. S-CNN \cite{shou2016temporal} adopts the "proposal + classification" paradigm to tackle this problem using multi-stage CNNs. SSN \cite{zhao2017temporal} designs a structured temporal pyramid architecture to hierarchically construct action proposals and then distinguish whether them are completed or not via two classifiers. Single Shot Action Detector (SSAD) \cite{lin2017single} draws on the method of object detection and detects action instances directly.
	
	\begin{figure*}[ht]
		\centering
		\includegraphics[width=0.95\linewidth]{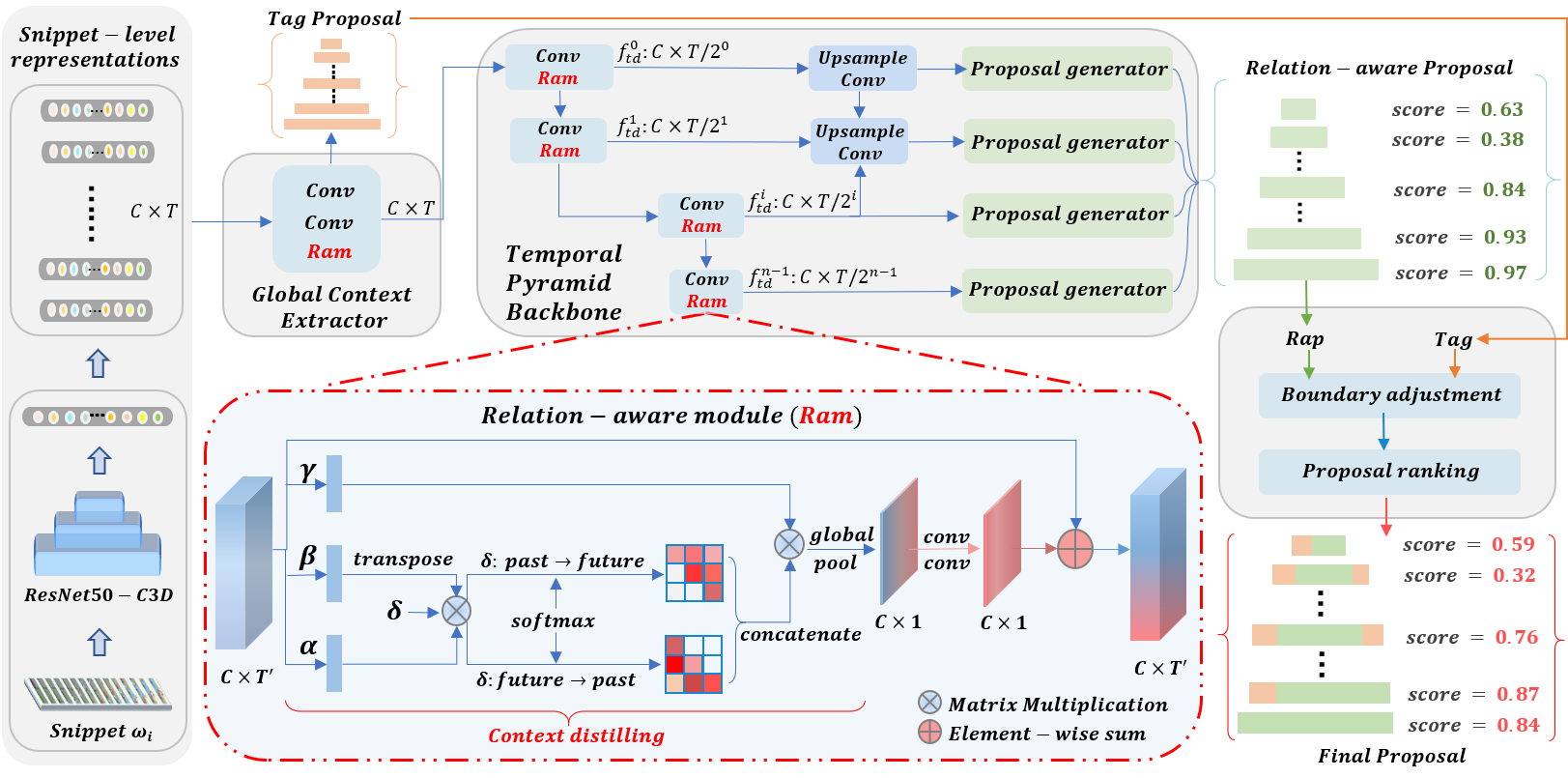}
		\caption{The framework of our approach. First, a ResNet50-C3D is used for snippet-level video representations. Then a global context extractor captures long-range dependency for predicting action probability on each snippet by actionness head. The following temporal pyramid backbone is enhanced with relation-aware modules for separately generating candidates instance with different duration via anchor head. Finally, our relation-aware proposals are combined with tag candidates to adjust boundaries and measure confidence score though proposal ranking for retrieving. }
		\label{fw}
	\end{figure*}
	\section{Relation-Aware Pyramid Network}
	Based on the above analysis, we propose a context distilling mechanism to capture long-range contexts in different temporal scales and combine them with corresponding localized contents for localizing action instance as accurately as possible.
	With this purpose in mind, we will introduce our Relation-Aware Pyramid network (RapNet), as shown in Fig.\ref{fw}. Specifically, we first introduce a novel relation-aware module (RAM) to extract the long-range context. Then a global context extractor employs a RAM to capture holistic information for augmenting full-scale local features. A following temporal pyramid backbone is also enhanced with our RAM to perform action proposals in multi temporal scales.
	
	\subsection{\emph {Relation-Aware Module}}
	In this section, we formulate how to distill long-range contexts over the entire video for augmenting the localized information in our relation-aware module, dotted box in Fig.\ref{fw}. This module first captures bi-directional relations between any two temporal positions as context distilling. Then, it squeezes temporal information to generate channel-wise mean statistics as the global context. Afterward, two fully-connected layers are introduced for adaptive feature responses recalibration.
	
	Given a video representation sequence $\Omega$, the content of each snippet is encoded into a vector of $C$ dimension so that $\Omega$ with a shape $C\times T$. Our relation-aware module can capture the long-range dependency between any two temporal location in $\Omega$ and have a matrix $M\in \mathbb{R}^{T\times T}$ shape.
	
	As discussed in \cite{wang2018non}, computing interactions between any two positions of features is an efficient, generic and straightforward way to capture global understanding of the visual scene. We follow this principle and define the relations between temporal positions as a dot-product affinity in the embedding spaces as:
	\begin{equation}
	\centering
	m(x_i, x_j) = \delta_{ij}\beta(x_i)^T\alpha(x_j),
	\label{relation}
	\end{equation}
	where $\alpha$, $\beta$ are two projection functions, which can be implemented by 1D convolution operation. $\delta_{ij}$ denotes the index function that is used to control whether the visual content of $j^{th}$ snippet has an impact on that of $i^{th}$.
	
	Then, we refine the features at $i^{th}$ snippet with visual contents of the $j^{th}$ one via matrix $M$ and we can have that:
	\begin{equation}
	\centering
	\hat{x_i} = \sum_{j=0}^{T}\gamma(x_i) m(x_i, x_j),
	\label{map_r}
	\end{equation}
	where $\gamma(\cdot)$ is an embedding function with learnable parameter $W_\gamma \in \mathbb{R}^{C\times C}$. After Eq.\ref{relation}, \ref{map_r}, the original features map at each temporal location can be refined by capturing contextual information of other temporal positions. Afterward, we propose to squeeze the temporal dimension to generate channel-wise mean statistics $X^G \in \mathbb{R}^C$, in order to obtain the global context embedding. Finally, we learn channel-wise interdependency for recalibrating feature responses and augment the original features $x_i$ with the global context so that:
	\begin{equation}
	\centering
	x^{'}_i = x_i + W_2\sigma(\varepsilon(W_1x^G_i)),
	\end{equation}
	where $W_2 \in \mathbb{R}^{C\times \frac{C}{r}}$ and $W_1 \in \mathbb{R}^{\frac{C}{r}\times C}$ with pre-defined reduction ratio $r$, $\sigma, \varepsilon$ is an activation function and normalization operation respectively.
	
	\textbf{Bi-directional relations.} Due to timing characteristics of video sequences, we not only need to capture the global contextual information, but also consider the role of directed relation between visual contents of two snippets when localizing activity instances in untrimmed videos. Hence, we can define directed relationship via our relation-aware module as following two forms, also shown in Fig.\ref{fw}
	
	\begin{minipage}{0.2\textwidth}
		\centering
		\begin{equation}
		\delta_{ij} =\begin{cases}
		0, & i \le j \\
		1, & i > j
		\end{cases},
		\label{tril}
		\end{equation}
	\end{minipage}
	\begin{minipage}{0.2\textwidth}
		\centering
		\begin{equation}
		\delta_{ij} =\begin{cases}
		0, & i > j \\
		1, & i \le j
		\end{cases}.
		\label{triu}
		\end{equation}
	\end{minipage}
	
	For Eq.\ref{tril}, the relationship matrix $M$ becomes an lower triangular one. Our relation-aware module only considers the contents of snippets behind current snippet $x_i$. Thus, the refined local pattern $\hat{x_i^+}$ can be defined as:
	\begin{equation}
	\centering
	\hat{x_i^+} = \sum_{j\ge i}\gamma(x_i) \alpha(x_i)^T\beta(x_j).
	\label{map_r_tril}
	\end{equation}
	Similarity, when $M$ follows the form in Eq.\ref{triu}, this upper triangular matrix indicates our relation-aware module can capture what happens in the past and learn its impact on the future. An enhanced feature $\hat{x_i}$ follows the format as:
	\begin{equation}
	\centering
	\hat{x_i^*} = \sum_{j\le i}\gamma(x_i) \alpha(x_i)^T\beta(x_j).
	\label{map_r_triu}
	\end{equation}
	Hence, Our relation-aware module models the bi-directional relation between two local patterns via a concatenation form defined as $\hat{x_i} = [\hat{x_i^+}, \hat{x_i^*}]$.

	\subsection{\emph{Architecture}}
	In this section, we introduce the details of our relation-aware pyramid network (RapNet) in Fig.\ref{fw}, which consists of three components: Global Contextual Extractor (\textbf{GCE}), Temporal Pyramid Backbone (\textbf{TPB}), Proposal Generator (\textbf{PG}). Previous works \cite{lin2017single,liu2019multi} also attempt feature pyramid networks for localizing action instances. However, they are unaware of the global contextual information and directed relations between two snippets. We propose relation-aware module to learn global context at different semantic levels for augmenting corresponding localized information when predicting actionness and proposals.
	
	\textbf{Global Contextual Extractor (GCE):}
	For a $T\times C$ video representation, two 1D temporal convolution layers, and one \emph{relation-aware module} serve as an extractor to capture the long-range context, which augments the feature used to predict action probabilities.
	
	\textbf{Temporal Pyramid Backbone (TPB):}
	Our temporal pyramid backbone uses $N$-level down-sampling operations to get a top-down feature pyramid set $F_{td}= \{ f^0_{td},\cdots, f^i_{td} ,\cdots, f^{N-1}_{td} \}$, each $i^{th}$ level feature $f^i$ with $\frac{T}{2^i}$ length of time in temporal scale $\frac{2^i}{T}$. Each operation consists of one \emph{relation-aware} module and two $Conv$-$BatchNorm$-$Relu$ blocks, which captures holistic information at current time scale to augment higher-level semantic features extracted at larger time scale.
	
	Aiming to produce multi-granularity temporal action proposals, we adopt lateral connections and up-sampling operations with a scale factor of 2 in the backbone. Relation-aware feature $f^i_{td}$ captured in $i^{th}$ level is passed to combine with the up-sampled feature $f_{up}^{i+1}$ from $i^{th}+1$ level. For this combination, high-level features are first reduced by half in channels and then up-sampled to twice the time scale, which are finally concatenated with the corresponding low-level ones. It is guaranteed that high-level semantic information and the global context in a large time scale are perceived by the lower-level ones. Hence, we have the bottom-up feature pyramid set for proposal prediction $F_{bu} = \{f^0_{bu},\cdots, f^i_{bu} ,\cdots, f^{N-1}_{bu} \}$.
	
	\textbf{Proposal Generator (PG):}
	We use a special prediction module, called proposal generator, to predict confidence scores, Interaction-over-Union (IoU) and location offsets of anchor-based action instances associated with each element in $F_{bu}$. The confidence score is used to indicate the probability of a prediction containing an action instance. The other two predictions help our RapNet localize a true action instance as accurately as possible. In addition, we employ IoU value as double-check for scoring a proposal prediction, which can correct the confidence error via multiplication. As shown in Fig.\ref{fw} with green boxes, these generators are implemented by two \emph{Conv-BatchNorm-Relu} blocks with kernel size 3, stride size 1. Every temporal scale feature $f^i_{bu}$ has its own generator to predict $M$ action instances at each temporal location so that generating $\frac{T}{2^i}\times M$ proposals in total. Hence, each generator will output $\frac{T}{2^i}\times M \times 4$ predictions, that is $P_i=\{p^0,\cdots, p^j,\cdots, p^{\frac{T}{2^i}}\}$ and $p^j=\{(\hat{p}^{jk}_{conf},\hat{p}^{jk}_{iou},\hat{p}^{jk}_{c},\hat{p}^{jk}_{w} )\}^M_{k=0}$. Since we have the location offsets $\hat{p}^{jk}_{c},\hat{p}^{jk}_{w}$, the adjusted location is defined as:
	\begin{equation}
	\centering
	\begin{aligned}
	\phi^{jk}_c = (t^{jk}_c + p^{jk}_c) / t^i_s \\
	\phi^{jk}_w = t^{jk}_w * exp^{p^{jk}_w}
	\end{aligned},
	\end{equation}
	where $t^{jk}_c$ and $t^{jk}_w$ are center and width of $k^{th}$ anchor instance at the $j^{th}$ temporal location in  $t^i_s = \frac{T}{2^i}$ time scale. It is easily to get the predicted action instance $\phi^{jk} = [\phi^{jk}_c - \frac{1}{2} \cdot \phi^{jk}_w, \phi^j_c + \frac{1}{2} \cdot \phi^j_w]$ and the total generations can be denoted as:
	$P = \{P^i=\{ p^j=\{(\hat{p}^{jk}_{conf},\hat{p}^{jk}_{iou},\hat{p}^{jk}_{c},\hat{p}^{jk}_{w} )\}^M_{k=0}\}^{\frac{T}{2^i}}_{j=0} \}^{N-1}_{i=0} $.
	
	\subsection{\emph{Loss Function}}
	Our RapNet captures the global contextual information at different temporal scales to produce anchor-based proposals and snippet-level actionness. Hence, the objective function for training can be formulated with a $L_2$ regularization as:
	\begin{equation}
	\centering
	L_{total} = \lambda_1 L_{prop} + \lambda_2 L_{action} + \lambda_3 L_2(W),
	\end{equation}
	where $L_{prop}$ and $L_{action}$ are the loss functions defined for proposal generation and actionness prediction respectively. We empirically set $\lambda_1 = 10 \lambda_2$, $\lambda_3=0.0005$ for experiments. To explain these loss functions in detail, we introduce the label assignment for these two parts.
	
	\textbf{Label assignment.} For anchor-based proposal prediction, we tag a binary label for each anchor instance, similar to YOLO \cite{redmon2016you}, that a positive label is assigned to the one with highest Interaction-over-Union (IoU) with corresponding ground-truth instance, otherwise negative. In this way, a ground truth instance only can match one anchor so that the boundary regression and IoU loss only considers positive samples. Due to this label assignment, it is imbalanced that the ratio between positive and negative training samples. Thus we adopt a screening strategy to ignore some negative instances for confidence loss. A negative instance will be ignored if the highest IoU overlap between ground-truth instances with all proposal predictions is larger than a threshold $\theta_{iou}$, empirically 0.5. For actionness generation, we follow the way in BSN \cite{lin2018bsn} and expand the action instance length by a ratio $\eta=0.1$. For a ground truth instance $\phi = [t_s, t_e]$, the label of each temporal location lying in the expanded region $[t_s - d \eta, t_e + d \eta]$ will be set to 1, where $d = t_e - t_s$.
	
	\textbf{Proposal generation losses.} Our RapNet predicts multiple bounding boxes per temporal grid cell. When training, we only want one anchor to be responsible for each action instance so that we only calculate the one which has the highest current IoU with the ground truth in label assignment. Hence, we optimize the following loss function $L_{prop}$:
	\begin{equation}
	\centering
	\begin{aligned}
	\lambda_{conf} [ \frac{1}{N_{pos}} \sum_{i=0}^{N-1}\sum_{j=0}^{\frac{T}{2^i}}\sum_{k=0}^{M}\Delta^{ins}_{ijk} f_{conf}(\hat{p}^{jk}_{conf},{p}^{jk}_{conf} ) \\
	+ \frac{1}{N_{neg}} \sum_{i=0}^{N-1}\sum_{j=0}^{\frac{T}{2^i}}\sum_{k=0}^{M}\Lambda^{ins}_{ijk} f_{conf}(\hat{p}^{jk}_{conf},{p}^{jk}_{conf} )] \\
	+ \lambda_c \frac{1}{N_{pos}} \sum_{i=0}^{N-1}\sum_{j=0}^{\frac{T}{2^i}}\sum_{k=0}^{M}\Delta^{ins}_{ijk} f_c(\hat{p}^{jk}_c, p^{jk}_c) \\
	+ \lambda_w \frac{1}{N_{pos}} \sum_{i=0}^{N-1}\sum_{j=0}^{\frac{T}{2^i}}\sum_{k=0}^{M}\Delta^{ins}_{ijk} f_w(\hat{p}^{jk}_w, p^{jk}_w) \\
	+ \lambda_{iou} \frac{1}{N_{pos}} \sum_{i=0}^{N-1}\sum_{j=0}^{\frac{T}{2^i}}\sum_{k=0}^{M}\Delta^{ins}_{ijk} (1 - iou_{jk}),
	\end{aligned}
	\label{prop_loss}
	\end{equation}
	where $N_{pos}$ and $N_{neg}$ represent the number of positive $\Delta^{ins}_{ijk}$ and screened negative $\Lambda^{ins}_{ijk}$ training instances respectively while ${p}^{jk}_{conf}$ is the binary label in label assignment, $f_{conf}(\cdot)$ and $f_c(\cdot)$ are binary cross-entropy with logits loss functions. $f_w(\cdot)$ represents smooth-$L_1$ loss and $iou_{jk}$ is the interaction-over-union between a prediction with ground truth. We set $\lambda_{conf}=0.2$ and other weights as 1 for training RapNet.
	
	\textbf{Actionness prediction losses.} Different from the way in BSN \cite{lin2018bsn}, we only consider the actionness score without the starting and ending probability prediction. A head layer is used to take the output of GCE module as input to predict action probabilities $p^i_a$ on each snippet $t_i$ of the entire video $T$. Our actionness generation also captures the long-range context to determine whether there is an action at the current temporal location. Hence, the objective function is defined as:
	\begin{equation}
	\centering
	L_{action} = \frac{1}{T}\sum_{i=1}^{T}( \Delta_i log(p^i_a) + (1 - \Delta_i) log(1- p^i_a)),
	\end{equation}
	where $T$ is the length of a video representation and $\Delta_i$ indicates the binary label assigned in label assignment.

	\subsection{\emph{Adjustment and Ranking}}
	Due to the complementary characteristics between anchor-based methods and actionness score based approaches, we design a two-stage scheme to adjust boundaries of our relation-aware proposals with tag candidates firstly and then measure the confidence score of refined proposals with corresponding action probabilities. Finally, we suppress the redundant proposals with Soft-NMS for retrieving \\
	\textbf{Boundary Adjustment} We first employ a TAG \cite{zhao2017temporal} technique to generate another set of proposals $P_{tag}$ for refining proposal boundaries. Specifically, for each candidate in $P_{rap}$ we calculate its IoU with all that in $P_{tag}$. If the maximum IoU is greater than a threshold, the boundaries of our relation-aware proposal $P_{rap}$ will be rectified by the corresponding candidate with a ratio $r$. Thus a refined candidate $\phi_{prop}^* = [\phi_{tag}^s * r + \phi_{prop}^s * (1-r), \phi_{tag}^e * r + \phi_{prop}^e * (1-r) ]$.\\
	\textbf{Proposal Ranking} We seek another mechanism to evaluate the confidence score here. This double-check scheme can bring improvements via a simple multilayer perception model enhanced with our relation-aware module. For a refined instance $\phi_{i}^*$ in $P_{rap}^*$, we first construct a proposal-level feature with fixed length and then estimate its overlap with ground truth directly. This predicted value multiply the original confidence score to serve as the final ranking score.\\
	\textbf{Proposal Suppression.} For a ground truth action instance, our RapNet will generate multiple candidates with different temporal IoU, which cause a low recall performance. Thus, we employ Soft-NMS \cite{bodla2017soft} to suppress redundant proposals with a score decaying function by a pre-defined threshold. For the ranked proposals, the score of candidate $\phi^*$ with the relatively low score will be decayed if its overlap with the proposal with a higher score is greater than the threshold.
	
	\begin{figure}[t]
		\centering
		\includegraphics[width=0.95\columnwidth]{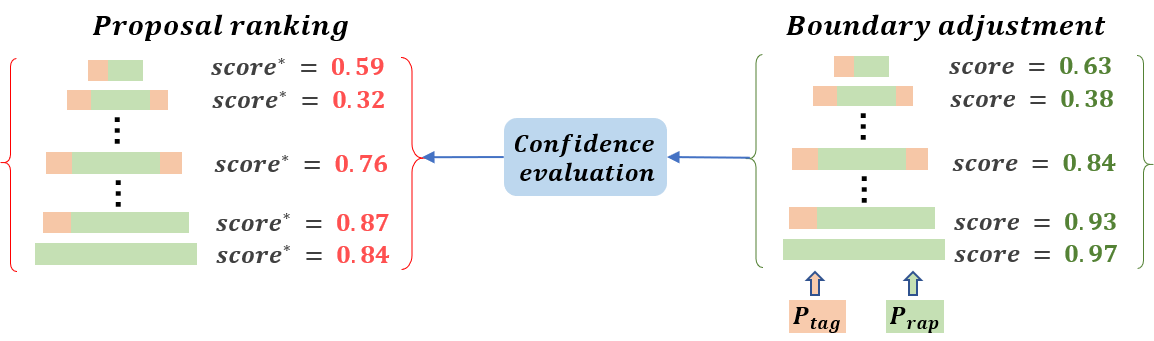}
		\caption{The scheme of boundary adjusting and proposal ranking}
		\label{adj_rank}
	\end{figure}
	
	\section{Experiments}
	\subsection{\emph{Datasets and Setup}}
	\textbf{THUMOS 2014.} This dataset consists of 13320 trimmed videos of 101 categories from UCF-101 for training, 1010 and 1574 untrimmed videos for validation and test set respectively. It is challenging, and widely used for temporal action detection task with 20 annotated sport categories. We conduct experiments the same as \cite{xiong2017pursuit}. \\
	\textbf{ActivityNet-v1.3.} It contains 19994 videos labeled in 200 classes. it is divided into training, validation and test with a ratio of 0.5, 0.25, 0.25 respectively. Due to the unavailability of annotations on test split, we compare and report performances of our approach and other state-of-the-art methods on the validation set, unless otherwise stated.\\
	\textbf{Video representations.} First, we extract RGB frames from the original video and then adopt an ResNet-50 C3D \cite{he2016deep,tran2015learning} to encode the visual context. In order to obtain compact features, we compose snippets sequence $\Omega = \{\omega_i\}_{i =1}^{T'} $ of a given video, where each snippet $\omega_i$ with $L$ frames and $T'$ is the number of snippets . Our Residual C3D network takes this sequence as input to generate encoded representations $f(\omega) \in \mathbb{R}^{T'\times C}$ of a given video. This net is first trained on Kinetics \cite{carreira2017quo} and then fine-tuned on ActivityNet-v1.3 and THUMOS14. Specifically, we add another fully connected layer for dimension reduction. Due to the different temporal length between videos, we resize each video encoding to a fixed temporal length so that we have a $T\times C$ feature map, \eg $T = 128, C = 256$ in ActivityNet-v1.3. \\
	\textbf{Implementation.} During generating snippet-level features, we set the clip interval as 16 and 5 for ActivityNet-v1.3 and THUMOS14 respectively. We also rescale the feature sequence of each video to a shape 128$\times$256 for the former. For proposal generation, we train our RapNet with batch size of 16 and initialize learning rate as 0.005 with Cosine Decline Learning Strategy for 18 epochs, which is warmed up four epochs in a linear growth mode. We set $N = 6$, $M=2$ for almost experiments, and also explore the effect of pyramid level $N$ and anchor number per location $M$ on the performance in ablation study. Different from previous works, we use K-means algorithm to cluster $N*M$ anchor boxes.\\
	\textbf{Evaluation metric.} We use the area under the average recall vs. average number of proposals per video curve (AUC), denoted as AR@AN. A proposal is a true positive if its temporal intersection over union (tIoU) with a ground-truth segment is not less than a given threshold (\eg, tIoU $\ge$ 0.5). AR is defined as the mean of all recall values using tIoU between 0.5 and 0.9(inclusive) in ActivityNet-v1.3, while between 0.5 and 1.0 in THUMOS 2014, with a step size of 0.05. AN is defined as the total number of proposals divided by the number of videos in the testing subset. Specifically, AN ranges from 0 to 100 in ActivityNet-v1.3.
	
	\begin{table}[t]
		\centering
		\caption{Comparison between our method with other state-of-the-art methods on ActivityNet-v1.3 in terms of AR@AN}
		\begin{tabular}{c|c|c|c}
			\toprule
			\midrule
			Method & AR@100 & AUC (val) & AUC (test) \\
			\midrule
			SSAD$^{\rm +}$ & 73.01& 64.40& 64.80 \\
			\midrule
			CTAP & 73.17 & 65.72 & - \\
			\midrule
			\multirow{2}{*}{BSN} & 74.16 & 66.17 & 66.26 \\
			\cline{2-4}
			&75.61{\rm *} & 67.47{\rm *} & 67.54{\rm *} \\
			\midrule
			MGG & 74.54 & 66.43 & 66.47 \\
			\midrule
			\multirow{2}{*}{RapNet} & 76.71 & 67.63 & 67.72 \\
			\cline{2-4}
			&\textbf{78.63{\rm *}} & \textbf{69.93{\rm *}} & \textbf{70.07{\rm *}} \\
			\midrule
			\bottomrule
		\end{tabular}
		\label{act_all}
		\footnotesize{\\$^{\rm +}$ the proposal generation part in \cite{lin2017single} \\ $^{\rm *}$ results are reported based on our ResNet50-C3D features}
	\end{table}

	\subsection{\emph{Temporal Proposal Generation}}
	In this section, we compare our relation-aware proposals generated from RapNet with other existing state-of-the-art methods on both ActivityNet-v1.3 and THUMOS14 datasets. For fair comparisons, we adopt the two-stream and C3D features provided by BSN \cite{lin2018bsn} to conduct experiments in considering that the effect of different feature representations. However, we use our C3D features, a better one, to exploit the ablation study in ActivityNet-v1.3 avoiding the insufficient video representations, which can measure the maximum performance of our RapNet.
	
	Tab. \ref{act_all} illustrates the AR@AN performances on the validation set of ActivityNet-v1.3 between our RapNet and other state-of-the-art methods, including MSRA \cite{yao2017msr}, Prop-SSAD\cite{lin2017single}, CTAP\cite{gao2018ctap}, BSN\cite{lin2018bsn}, MGG\cite{liu2019multi}. No matter in terms of AUC or AR@100, it is observed that our RapNet outperforms all of them. With the same feature, relation-aware proposal achieves 67.72 in test, around 1.3 higher than the state-of-the-art method MGG. Specifically, our approach improves the AR@100 on validation set from 75.61 of BSN to 78.63 with our ResNet50-C3D video representations, while the improvement is smaller when using the same two-stream features as BSN. It indicates that our RapNet is more suitable for C3D features.
	
	Tab. \ref{thumos_all} explicates the performance comparison on the testing set of THUMOS14, where we adopt the same representations as BSN. Our approach consistently improves the AR@AN performances from 50 to 1000 in both C3D features and 2-stream features. These results demonstrate the effectiveness of our RapNet for accurate temporal action proposal generation method, especially for the small average number of proposals per video.

	\begin{table}[t]
		\scriptsize
		\centering
		\caption{Comparison between our RapNet with other state-of-the-art proposal generation methods on THUMOS14}
		\begin{tabular}{ccccccc}
			\toprule
			\toprule
			Feature & Method & @50 & @100 & @200 & @500 & @1000 \\
			\midrule
			C3D & DAPs & 13.56 & 23.83 & 33.96 & 49.29& 57.64 \\
			C3D & SCNN & 17.22& 26.17 &37.01& 51.57& 58.20\\
			C3D & SST & 19.90 & 28.36 &37.90 &51.58 &60.27\\
			C3D & TURN & 19.63 & 27.96 &38.34 &53.52 &60.75 \\
			C3D & BSN & 29.58 & 37.38 & 45.55 & 54.67 & 59.48\\
			C3D & MGG & 29.11 & 36.31 & 44.32 & 54.95 & 60.98 \\
			\midrule
			C3D & RapNet & \textbf{29.72}& \textbf{37.53}& \textbf{45.61}&\textbf{55.26} & \textbf{61.32}\\
			\bottomrule
			\toprule
			Flow & TURN & 21.86 & 31.89 & 43.02 & 57.63 & 64.17\\
			2-Stream & TAG & 18.55 & 29.00 & 39.61 & - & - \\
			2-stream & CTAP & 32.49 & 42.61 & 51.97 & - & - \\
			2-Stream & BSN & 37.46 & 46.06 & 53.21 & 60.64 & \textbf{64.52} \\
			2-Stream & MGG & 39.93 & 47.75 & 54.65 & 61.36 & 64.06 \\
			\midrule
			2-Stream & RapNet & \textbf{40.35}& \textbf{48.23}& \textbf{54.92}& \textbf{61.41}& 64.47 \\
			\bottomrule	
			\bottomrule
		\end{tabular}
		\label{thumos_all}
	\end{table}
	
	\subsection{\emph{Ablation Study}}
	
	\textbf{How deep?} We conduct experiments to exploit the performance with different depths of pyramid network. First, we fix the anchor number $M=2$ and generate candidates only via the pyramid network without relation-aware module and actionness score prediction. Finally, we evaluate its performance on the ActivityNet-v1.3 validation set in terms of AR@AN. The detailed results are shown in Tab.\ref{explore_depth}. It is observed that the AUC is increased sharply as the network deepens, while the gain effect is getting smaller. When the depth $N$ increases from 3 to 4, it improves the AUC by around 2.5, but this improvement decreases to 0.2 when $N$ from 5 to 6. It indicates the effectiveness and necessity of detecting long action instances and short ones at different temporal scales.
	\begin{table}[t]
		\centering
		\caption{Comparison performance on ActivityNet-v1.3 in terms of AR@AN with different $N$, reported in percentage}
		\begin{tabular}{c|c|c|c|c|c}
			\toprule
			\midrule
			Depth & AUC & @1 & @5 & @10 & @100 \\
			\midrule
			3 & 64.54 & 23.79& 42.52& 50.70& 74.28\\
			4 &  67.07& 30.07 & 46.06 & 54.02 & 76.43\\
			5 & 68.16 & 33.55 &  48.32& 55.67 & \textbf{77.18}\\
			6 & \textbf{68.35} & \textbf{33.55} & \textbf{48.48} & \textbf{55.79} &  77.01\\
			\midrule
			\bottomrule
		\end{tabular}
		\label{explore_depth}
	\end{table}
	
	\begin{table}[t]
		\caption{Comparison with different part of our relation-aware module on ActivityNet-1.3 in terms of AR@AN}
		\resizebox{.95\columnwidth}{!}{
		\begin{tabular}{c|c|c|c|c|c}
			\toprule
			\midrule
			Module & AUC & @1 & @5 & @10 & @100 \\
			\midrule
			TPB & 68.35 & 33.55 & 48.48 & 55.79 &  77.01\\
			TPB{\rm *} & 69.47& 33.69& 48.97& 56.70 & 78.07 \\
			GCE + TPB{\rm *} & 69.53& 34.32& 49.43& 56.52& 78.08\\
			\midrule
			\bottomrule
		\end{tabular}
		}
		\footnotesize{\\$^{\rm *}$ The relation-aware-module enhanced temporal pyramid network}
		\label{explore_rab}
	\end{table}
	\textbf{Why relation-aware module?} We generate proposals using a pyramid network with and without relation-aware module (RAM) respectively for exploiting the effect of context distilling. In Table \ref{explore_rab}, the top row is the result (68.35\% in AUC) of plain pyramid net without our RAM and actionness prediction. The middle row is the performance of pyramid net enhanced with our RAM. It achieves 69.47\% in AUC, 1.1 higher than the plain counterpart. It demonstrates the ability that relation-aware modules capture the global context to augment local features for higher quality proposals. Especially, for large AN, \eg @10 or @100, the RAM helps neural networks to focus on informative content rather than background. Compared the last two rows, it displays that additionally actionness prediction benefits the proposal generation. When considering a few candidates per video, such as AR@1 or AR@5 in Tab. \ref{explore_rab}, it is concluded that the RapNet has better confidence score than its counterpart without actionness branch.
	
\begin{figure}
		\centering
		\includegraphics[width=0.95\columnwidth]{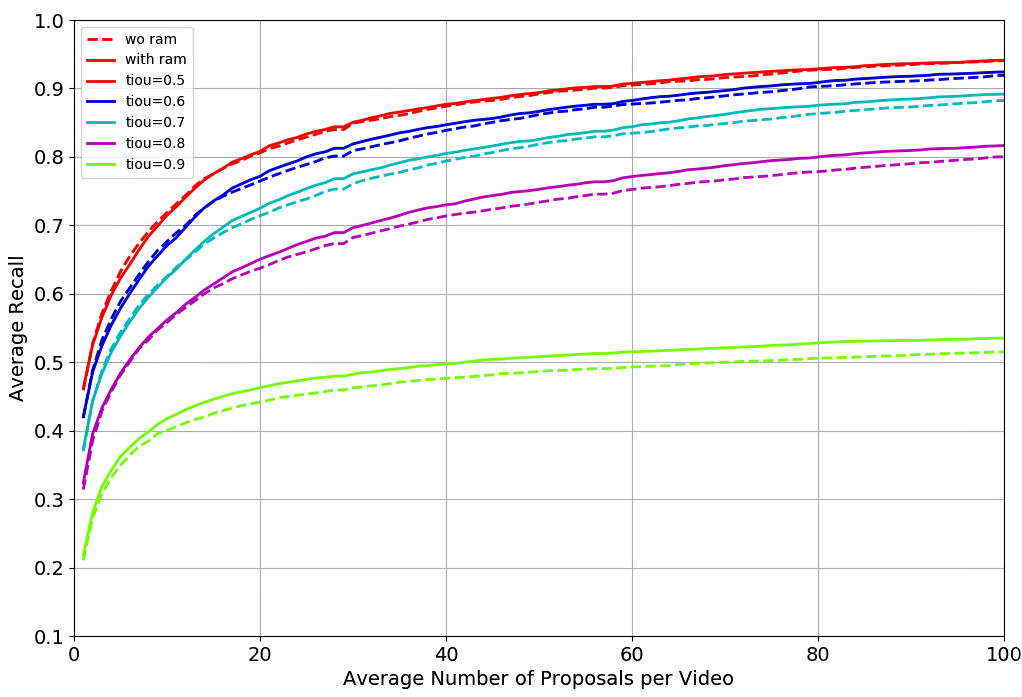}
		\caption{Comparisons on pyramid net with or wo relation-aware module in terms of AR-AN curves of different tIoU }
		\label{fig_tiou}
	\end{figure}

	Illustrated in Fig.\ref{fig_tiou}, we explore the boundary precision between plain pyramid net and the relation-aware enhanced one on different tIoU threshold. The proposed module improves average recall by a great margin, especially in higher tIoU. It can be attributed to more precise boundaries of proposals benefited from capturing global context at temporal multi-scales to augment the localized information.
	
	\begin{table}[t]
		\centering
		\caption{Comparison with different numbers of anchor box in our RapNet on ActivityNet-1.3 in terms of AR@AN}
		\begin{tabular}{c|c|c|c|c|c}
			\toprule
			\midrule
			Numbers & AUC & @1 & @5 & @10 & @100 \\
			\midrule
			6 & 68.02 &33.75 & 49.64& \textbf{57.62}& 75.41\\
			12 & \textbf{69.53}& \textbf{34.32}& 49.43& 56.52& 78.08 \\
			18 & 69.24 & 33.61 & \textbf{49.64} & 56.94 & 77.81\\
			24 & 69.19 & 33.88 & 49.42 & 56.20 & \textbf{78.55} \\
			\midrule
			\bottomrule
		\end{tabular}
		\label{explore_anchor}
	\end{table}
	\textbf{How many anchor boxes?} We fix the depth of pyramid net as six and cluster different number ($M=1,2,3,4$) of anchors for analyzing the effect of these predefined anchor boxes. Table \ref{explore_anchor} displays the comparison between different anchors on the validation set of ActivityNet-v1.3 in terms of AR@AN. When $M=2$, RapNet achieves the best performance in AUC. Combined with the results in Tab.\ref{explore_depth}, it is found that the corresponding performance improvement gets limited when the total number of anchor is larger than 10.
	
	\textbf{Why proposal refinement?} We explore the two-stage scheme to refine generated proposals via boundary adjustment and ranking. From Table \ref{explore_adj_rank}, the gain effect of boundary adjustment and proposal ranking is 0.18 and 0.22, respectively. Compared the considerable performance improvement in previous works\cite{gao2018ctap,liu2019multi}, it reveals that RapNet enhances the segmental proposal generation and narrows the gap between sliding-window based and actionness score based approaches.

	\begin{table}[th]
		\caption{Performance of two-stage boundary adjustment and proposal ranking scheme on ActivityNet-1.3}
		\begin{tabular}{c|c|c|c|c|c}
			\toprule
			\midrule
			Methods & AUC & @1 & @5 & @10 & @100 \\
			\midrule
			RapNet & 69.53& 34.32 & 49.43 & 56.52 & 78.08 \\
			+ adjustment & 69.71 & 34.61& 50.24& 57.66& 78.60\\
			+ ranking$^{\rm *}$ & 69.93& 34.96& 50.27& 57.43& 78.63\\
			\midrule
			\bottomrule
		\end{tabular}
		\footnotesize{\\$^{\rm *}$ ranking the proposals with refined boundaries}
		\label{explore_adj_rank}
	\end{table}
	
	\section{Conclusion}
	In this work, a novel relation-aware-pyramid network is proposed for temporal proposal generation. We introduce a relation-aware module to captures long-range multi-scale context to augment local features for extracting informative contents and suppressing the backgrounds. A pyramid network enhanced with our relation-aware module has achieved state-of-the-art performance than other competitive methods on ActivityNet-v1.3 and THUMOS14. However, the boundaries of the generated proposal and the scoring mechanism are expected to be improved in future work.
	
	\section{Acknowledgement}This work was performed when Jialin Gao was an intern in CloudWalk Technology Co., Ltd. It was also partially supported by grant from the National Natural Science Foundation of China (61772513). Shiming Ge is also supported by the Youth Innovation Promotion Association, Chinese Academy of Sciences.
	
	\bibliographystyle {aaai}
	\bibliography {1378-bib}

\end{document}